\documentclass{article}
\usepackage[T1]{fontenc} 
\usepackage[utf8]{inputenc} 
\usepackage{ismir,amsmath,cite,url}
\usepackage{graphicx}
\usepackage{color}
\usepackage{amsmath,amsfonts}
\usepackage{booktabs}
\usepackage{placeins}

\DeclareMathOperator*{\argmax}{arg\,max}

\title{Convolutional Composer Classification}

\multauthor
{Harsh Verma \hspace{1cm} John Thickstun} {
  Allen School of Computer Science and Engineering,\\ University of Washington, Seattle, WA, USA\\
{\tt\small harshv@cs.washington.edu, thickstn@cs.washington.edu}
}
\def\authorname{Harsh Verma, John Thickstun}

\begin{document}

\maketitle
\begin{abstract}
This paper investigates end-to-end learnable models for attributing composers to musical scores. We introduce several pooled, convolutional architectures for this task and draw connections between our approach and classical learning approaches based on global and n-gram features. We evaluate models on a corpus of 2,500 scores from the KernScores collection, authored by a variety of composers spanning the Renaissance era to the early 20th century. This corpus has substantial overlap with the corpora used in several previous, smaller studies; we compare our results on subsets of the corpus to these previous works.
\end{abstract}

\section{Introduction}\label{sec:intro}

Models for attributing composers to musical scores have been extensively studied in the music information retrieval community. The composer classification question has been posed for a variety of corpora, from Renaissance composers \cite{buzzanca2002supervised,brinkman2016musical}, to the narrow (and challenging) case of Haydn and Mozart string quartets \cite{hillewaere2010string,herlands2014machine,van2005musical,kempfert2018does}, and to various collections of classical era composers (most of the other papers discussed in Section \ref{sec:rel}). In this work we study an expansive collection of scores, from 13th century sacred music by Guillaume Du Fay to 20th century ragtimes by Scott Joplin. 

A major challenge of this task is learning from limited data. While the corpus considered here is larger than most, this is largely due to the number of composers considered (19): for specific composers, we have at most 466 scores (Bach) and as few as 22 (Japart). Small datasets are an inherent problem for composer classification: the corpus used in this work contains, for example, all of the Bach chorales and all of the Mozart string quartets. We cannot resurrect these composers and have them write us more scores to include in our corpus. This situation contrasts starkly with many learning problems, where substantial progress can be made by collecting massive datasets and exhaustively training an expressive model (usually a deep neural network) with ``big data.''

Further complicating this task, an individual score is itself a high dimensional object: the average score in our corpus consists of thousands of notes, each of which is encoded as a high dimensional vector to represent its pitch and value. Learning from a small number of examples in a high dimensional space is a formidable problem; thus much work on composer classification focuses on feature engineering, feature selection, dimensionality reduction, or some combination of these approaches to construct low-dimensional representations of scores to learn from.

In this paper we take a different approach: we dispense with feature engineering and explore end-to-end classifiers that operate directly on full scores. Specifically, we investigate shallow convolutional neural networks with an aggressive pooling operation. In this setting, all but the most impoverished linear classifiers achieve 100\% training accuracy. We rely on implicit regularization introduced by the network structure and first-order optimization with early stopping to avoid overfitting to training data. While theoretical understanding of such an approach is in its early stages \cite{soudry2018implicit}, we find empirically that this works quite well for composer classification.

\section{Related Work}\label{sec:rel}

The earliest works on composer classification \cite{pollastri2001classification,buzzanca2002supervised} analyzed highly preprocessed corpora of melodic fragments. Much of the subsequent work on classification focuses on engineering features to summarize full scores. These approaches can be broadly categorized, using the terminology of \cite{hillewaere2009global}, into ``global'' summarization approaches that compute small sets of summary statistics as a feature set for each score\cite{van2005musical,kaliakatsos2010musical,herlands2014machine,herremans2016composer,sadeghian2017classification,kempfert2018does,brinkman2016musical} and local ``event'' featurizations that extract n-gram counts of a score as features\cite{wolkowicz2008n,hillewaere2010string,kaliakatsos2011weighted,hontanilla2013modeling,wolkowicz2013evaluation}. There is also a line of work that applies compression-based dissimilarity metrics \cite{anan2012polyphonic,takamoto2016improving,takamoto2018feature} to this task, which offers a substantially different perspective on classification problems.

The present work is most similar in spirit to \cite{buzzanca2002supervised} and \cite{velarde2016composer}. Like \cite{buzzanca2002supervised}, we adopt an end-to-end approach to feature learning using neural architectures. In contrast with \cite{buzzanca2002supervised}, we learn on full scores with minimal preprocessing and consider a multi-class classification task over a broad variety of composers; this approach is made possible by modern hardware unavailable to researchers in 2002. We also take a more systematic approach to architecture exploration, and identify effective architectures that are simpler than hybrid convolutional-recurrent approach taken in \cite{buzzanca2002supervised}.

Like \cite{velarde2016composer}, we exploit structure in musical scores using convolutional models. But where \cite{velarde2016composer} use a fixed Morlet or Gaussian convolution filters, the convolutional filters in this work are parameterized and learned from the data to maximize classification accuracy. We also explore multi-layer ``deep'' convolutional models and demonstrate improvements using such architectures versus the single layer of convolutions explored in \cite{velarde2016composer}.

Comparing to the substantial body of work that emphasize feature-engineering, the present work can be seen as an unified framework for learning global and event features. We will draw analogies between linear convolutional filters and n-gram features, and also demonstrate how convolutional models can express many popular global features. We will also introduce a global pooling operation that can be interpreted as an counter that tracks the number of occurrences of learned features, which is directly analogous to the count and ratio statistics that comprise the bulk of metrics used in human-engineered featurizations.

\section{Corpus and Data Representation}\label{sec:data}

We train and evaluate models on a corpus of 2,500 scores spanning five centuries of choral, piano, and chamber compositions from the KernScores collection \cite{sapp2005online}. An overview of this collection is provided in Table \ref{fig:corpus}. In this work, we consider each movement of a multi-movement composition to be a distinct score. Our models extract only the note data (pitch, note-value, and voicing) from scores, ignoring all other markings such as time signatures, key signatures, tempo markings, instrumentation, and movement names. For the Renaissance composers in this collection (Du Fay through Japart) we shorten the length of all note-values by a factor of 4 to crudely account for the shift in duration conventions between mensural and modern notation \cite{cumming2000motet}.

{\tiny
\begin{table}[t!]
  \centering
  \setlength\tabcolsep{2pt}
  \begin{tabular}{lclr}
    \toprule
    \cmidrule{1-4}
    Composer & Dates & Sub-Collection & Scores  \\
    \midrule
    Du Fay & 1397-1474 & Choral & 35 \\
    Ockeghem & 1410-1497 & Choral & 98 \\
    Busnois & 1430-1492 & Choral & 68 \\
    Martini & 1440-1497 & Choral & 122 \\
    Compere & 1445-1518 & Choral & 27 \\
    Josquin & 1450-1521 & Choral & 423 \\
    de la Rue & 1452-1518 & Choral & 178 \\
    Orto & 1460-1529 & Choral & 43 \\
    Japart & 1474-1507 & Choral & 22 \\
    Corelli & 1653-1713 & Trio Sonatas & 188 \\
    Vivaldi & 1678-1741 & Concertos & 33 \\
    Bach & 1685-1750 & Chorales & 370 \\
     &  & Well-Tempered Clavier & 96 \\
    D. Scarlatti & 1685-1757 & Keyboard Sonatas & 59 \\
    Haydn & 1732-1809 & String Quartets & 209 \\
    Mozart & 1756-1791 & Piano Sonatas & 69 \\
     & & String Quartets & 82 \\
    Beethoven & 1770-1827 & Piano Sonatas & 102 \\
     & & String Quartets & 67 \\
    Hummel & 1778-1837 & Preludes & 24 \\
    Chopin & 1810-1849 & Preludes and Mazurkas & 76 \\
    Joplin & 1868-1917 & Ragtimes & 47 \\
    \bottomrule
  \end{tabular}
  \caption{Details of the KernScores collection used for training and evaluation in this paper. } 
 \label{fig:corpus}
 \vspace*{-4mm}
\end{table}
}

We represent a score by lossless encoding of its pitch, voice, and note-value contents, transcoded from a **kern file to a binary representation suitable for input to a neural network. Specifically, we encode a score $S$ as a binary tensor $\textbf{x} \in \mathcal{S} = \{0,1\}^{T\times P \times (N + D+1)}$ where $T,P,N,D$ are defined as follows:
\begin{itemize}
\setlength\itemsep{0em}
\item $T$ - The number of rows of pitch/note-value data in the score $S$.
\item $P$ - The maximum number of concurrent **kern columns (spines): 6 for this corpus.
\item $N$ - The range of note pitches: 78 for this corpus, ranging from C1 to F\#7.
\item $D$ - The number of distinct note values (i.e. durations): 55 for this corpus.
\end{itemize}
For each $t \in \{0,\dots,T-1\}$, $p \in \{0,\dots,P-1\}$, $n \in \{0,\dots,N-1\}$, and $d \in \{0,\dots,D-1\}$ we set
\begin{align*}\label{eqn:scoredef}
& \textbf{x}_{t,p,n} = 1 & & \text{iff pitch $n$ occurs at time $t$ in spine $p$},\\
& \textbf{x}_{t,p,N+d} = 1 & & \text{iff note-value $d$ occurs at time $t$ in spine $p$},\\
& \textbf{x}_{t,p,N+D} = 1 & & \text{iff pitch $n$ continues at time $t$ in spine $p$}.
\end{align*}

\begin{figure*}[t]
\begin{minipage}{0.5\textwidth}
\begin{align*}
& \texttt{23:} & &\texttt{2..r~~~~~~2..r~~~~~~2..r~~~~~~2..r}\\
& \texttt{24:} & &\texttt{8r~~~~~~~~8r~~~~~~~~8r~~~~~~~~8dd}\\
& \texttt{25:} & &\texttt{=1~~~~~~~~=1~~~~~~~~=1~~~~~~~~=1}\\
& \texttt{26:} & &\texttt{1r~~~~~~~~1r~~~~~~~~8r~~~~~~~~4dd}\\
& \texttt{27:} & &\texttt{.~~~~~~~~~.~~~~~~~~~8f\# 8a~~~~.}\\
& \textbf{{\color{red} \texttt{28:}}} & &\textbf{{\color{red} \texttt{.~~~~~~~~~.~~~~~~~~~8a 8f\#~~~~8ff\#}}}\\
& \texttt{29:} & &\texttt{.~~~~~~~~~.~~~~~~~~~8a 8f\#~~~~16ee}\\
& \texttt{30:} & &\texttt{.~~~~~~~~~.~~~~~~~~~.~~~~~~~~~16dd}\\
\end{align*}
\end{minipage}
\hspace{7mm}
\begin{minipage}{0.5\textwidth}
\includegraphics[width=.8\linewidth,scale=1.0]{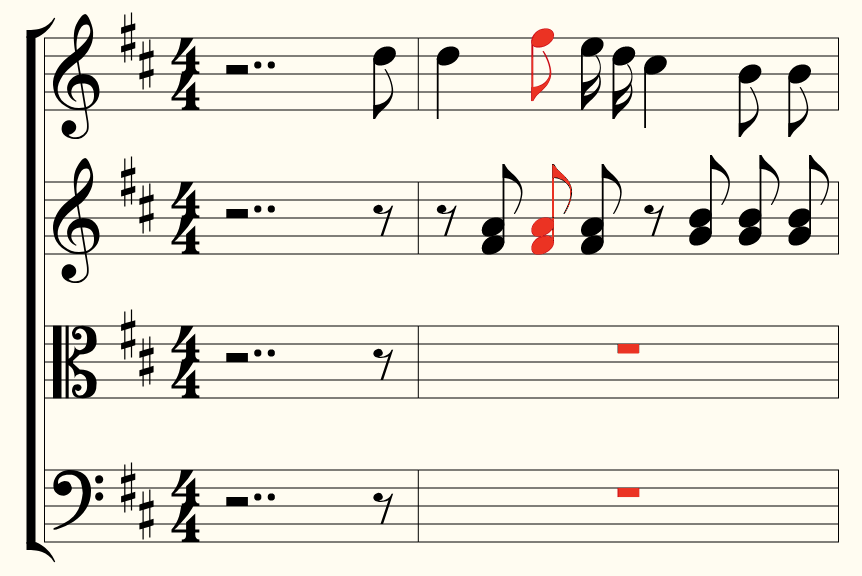}
\end{minipage}
\vspace*{-3mm}
\caption{Left: An excerpt from a **kern encoding of of Haydn's Opus 33, No 1, consisting of the first 6 beats of the first movement. Right: A visual rendering of the first two measures of the same score as sheet music. In the **kern format, time proceeds from top to bottom, whereas in traditional notation time proceeds left to right. The contents of the beginning of the 6th beat is highlighted (red) in each format to aid comparison between the **kern format and sheet music.}\label{fig:kernscore}
\vspace*{-3mm}
\end{figure*}

For example, consider how we would encode line 28 of the **kern excerpt shown in Figure \ref{fig:kernscore}. This is the 5th row of pitch/note-value data in the score, so we will encode data from this line into $\textbf{x}_5$. The periods ``\texttt{.}'' in columns (i.e. spines) 0 and 1 indicate that a note or (in this case) a rest is continued from a previous line, so we set $\textbf{x}_{5,0,N+D} = 1$ and $\textbf{x}_{5,1,N+D} = 1$. Two notes are indicated in spine 2: 8a and 8f. The number ``8'' indicates an eighth-note value. Each unique note-value is assigned an (arbitrary) index; we assign index 15 to the eighth-note value, so we set $\textbf{x}_{5,2,N+15} = 1$. The letters ``a'' and ``f'' indicate the pitches A3 and F4, which lie 33 and 41 semitones above C1 (the base of our note range) respectively. Therefore we set $\textbf{x}_{5,2,33} = 1$ and $\textbf{x}_{5,2,41} = 1$. Finally, spine 3 indicates an eighth-note F5 so we set $\textbf{x}_{5,3,53} = 1$ and $\textbf{x}_{5,3,N+15} = 1$.

The encoding defined about is an essentially verbatim transcoding of the **kern text data to a binary structured format. Converting from text to this structured format will allow us to write convolution operations along the time and pitch axes of the data tensor \textbf{x}. Encoding pitches with binary indicators in $N$-dimensional vectors is consistent with piano roll representations \cite{velarde2016composer} but departs from the example of \cite{buzzanca2002supervised}, which encodes pitch as a single numerical magnitude. The binary pitch encoding is required to support convolutions along the pitch domain, which we will introduce later in models \eqref{eqn:harmony} and \eqref{eqn:hybrid}.

The binary note-value encoding also differs from the numerical magnitude encoding used in \cite{buzzanca2002supervised}. We will not introduce models that convolve over durations (there is no translation invariant structure to exploit) so the motivation above for representing pitches with indicators does not apply to durations. Rather, we are motivated by the observation that note-values of similar duration may not be more alike in any musical sense than note-values with less similar duration. We avoid imposing this notion of similarity a-priori by encoding durations as categorical indicators: in this encoding, all note-values are equally distant in the Euclidean sense.

We also contrast our note-value encodings with piano roll representations, such as the representation used in\cite{velarde2016composer}. In a piano roll representation, time is discretized: the value of a note is indicated implicitly by the number of discrete time-slices over which it is sustained. We choose an explicit representation of note-values because it more directly reflects the contents of a written score, and results in shorter time series overall than discretized representations.

\vspace*{-2mm}
\section{Problem Formulation}\label{sec:task}

Our aim is to learn a classifier that predicts a composer $y$ given a score $\textbf{x}$. There are $C \equiv 19$ composers in our corpus: we assign each composer a label from 0 to $C-1$. We will construct a model $f_\theta : \mathcal{S} \to \{0,1\}^C$ that assigns vector $f_\theta(\textbf{x})$ to a score \textbf{x} where each component $f_\theta(\textbf{x})_i$ indicates model's (un-normalized) confidence that composer $i$ wrote score $\textbf{x}$. We predict $\hat y_\theta(\textbf{x}) \equiv \argmax_i f_\theta(\textbf{x})_i$, the composer the model has most confidence in. 

We evaluate our models via accuracy on holdout sets $\textbf{x}^{\text{holdout}}$, where accuracy is the zero-one loss defined by
\begin{equation*}
\text{Accuracy}(\textbf{x}^{\text{holdout}}) = \frac{1}{n}\sum_{i=1}^n \textbf{1}(\hat y_\theta(\textbf{x}_i^{\text{holdout}}) = y_i).
\end{equation*}
Here $\textbf{1} : \text{Bool} \to \{0,1\}$ is the indicator function: $\textbf{1}(p) = 1$ if the proposition $p$ is true, otherwise $\textbf{1}(p) = 0$. The results in Section \ref{sec:results} report 10-fold cross validated accuracies. It is standard practice in the machine learning community to report results on a single holdout set. But for for the small datasets considered in composer classification, cross-validating is essential to cut down the variance of estimated accuracy.

Given a collection of labeled scores (training data) $\{(\textbf{x}_1,y_1),\dots,(\textbf{x}_n,y_n)\}$ and a parameterized family of models $\{f_\theta : \theta \in \Theta\}$ we learn an optimal model $f_\theta$ by empirical risk minimization of the negative log-likelihood under the softmax-normalized probability distribution of model outputs:
\begin{equation}\label{eqn:obj}
\min_{\theta \in \Theta} \sum_{i=1}^n -\log \left(\frac{\exp(f_\theta(\textbf{x}_i)_{y_i})}{\sum_{k=1}^C \exp(f_\theta(\textbf{x}_i)_{y_k})}\right).
\end{equation}
For each of iteration of 10-fold cross-validation, in addition to the holdout fold $\textbf{x}^{\text{holdout}}$, we hold out a second fold as validation data and optimize the objective \eqref{eqn:obj} on the remaining 8 folds. We train our models using the Adam optimizer \cite{kingma2014adam}, regularizing with retrospective early stopping at the point with best accuracy on the validation fold.

\vspace*{-2mm}
\section{Models}\label{sec:models}

Every model class $f_\theta$ that we consider in this paper takes the following general approach.
\begin{enumerate}
\item Compute a set of local features at or around each time index in the score. 
\item Average these features across time (``pool'' the features, in neural networks parlance) into a single global feature vector.
\item Construct a linear classifier on top of this global feature vector to predict the composer of the score.
\end{enumerate}

The approach above is motivated by the need to manage the high-dimensionality of a score: given even the first 5 indices of the score tensor \textbf{x} described in Section \ref{sec:data}, we can easily fit a classifier that achieves 100\% training accuracy but fails to generalize to new data. As discussed in Section \ref{sec:rel}, the classical approach to this overfitting phenomenon is to reduce a score to a low-dimensional summary of pre-determined features and fit a classifier to this summary. The present work aims to learn features from scratch, but if we permit our model to learn any features it wants then it will simply overfit to the training data.

We therefore cripple our models in two ways. First, step~1 of the general approach above limits our model to learn features that are local in scope. We will allow our models to learn features that are sensitive to correlations between co-occurring notes (harmony), or between short sequences of notes (melody, rhythm). But by construction, our models will not be able to learn features that capture correlations between (for example) the first and last notes of a score. This precludes us from learning certain high-level patterns that could have predictive power (e.g. Mozart is more likely to repeat a section verbatim than Bach) but saves of us from learning a multitude of spurious patterns that appear to have predictive power on the training data but fail to generalize to new observations.

Second, step 2 of the general approach prevents our model from learning features that occur at fixed time locations. As discussed above, even knowing the first 5 indices of the score tensor is enough to easily identify every score in the corpus. By pooling features together across time, we force our models to classify based on the overall prevalence of the features it learns, rather than the occurrence of a particular feature at a particular time.

Note that classical approaches to feature engineering largely follow the same modeling restrictions outlined above. The engineered features used in e.g. \cite{herremans2016composer} (Table 1, page 7) or \cite{brinkman2016musical} (Table 1, page 2) consist primarily of overall frequencies, prevalences, and rates of occurrence. These features capture properties of either a single time index or short sequences, aggregated across an entire score. The use of n-gram features also fits this mold: an n-gram is by definition a local feature of n time indices, and an n-gram featurization computes aggregate (i.e. pooled) counts of the occurrences of each particular n-gram across a score. The use of genuinely non-local features is rare. They are used in \cite{kempfert2018does}: see for example the ``maximum fraction of overlap with opening material within first half of movement'' feature. The use of these features may account for the effectiveness of \cite{kempfert2018does} in the Haydn versus Mozart classification task, which our models underperform on.

\vspace{-2mm}
\subsection{Sub-sampling Scores}\label{subsec:subsample}

The approach outlined above requires us to average features across entire scores.  Each score in our corpus has a unique length, ranging from 10 to 4000 time indices. As a practical matter, it is difficult to deal with such variable-length data in machine-learning systems; our tools are designed to operate efficiently on homogeneous batches of data. One solution to this problem is to sub-sample scores; for example, \cite{velarde2016composer} train models on the first $s$ quarter notes where $s=70$ or $s=400$. Those authors found that the larger sample consistently outperforms the smaller one. We confirm this finding with the experiments in Table \ref{fig:contextlength}, which show that our models consistently perform better with larger samples of the score.

We therefore make the following compromise between using all available information from a score and operating on homogeneous inputs: we sample the first $s$, middle $s$, and last $s$ indices from our score $\textbf{x}$, resulting in $3s$ time indices sampled from each score. We use $s=500$ for all experiments except the experiments in Table \ref{fig:contextlength} that explore how models behave as we vary this hyperparameter. The average score in our corpus has 534 time indices, so for most scores this means we sample the entire score (for scores shorter than 500 time indices we pad out our sample with zeros). Only for scores longer than $1,500$ time indices (there are 117 in our corpus) do we lose any information with this approach.

\begin{table}[!t]
\begin{center}
  \setlength\tabcolsep{4pt}
  \begin{tabular}{ | l || c | c | c | c | c | c |}
    \hline
    & \multicolumn{6}{ c }{Sample Size} \vline \\ \hline
    Model & 10 & 20 & 50 & 100 & 250 & 500 \\ \hline\hline
    Histogram (Eqn \ref{eqn:hist}) & 50.0 & 59.0 & 62.0 & 63.0  & 66.1 & 64.2 \\ \hline
    Voices (Eqn \ref{eqn:partdeep}) & 60.0 & 61.6 & 63.9 & 72.0 & 75.5 & 76.9 \\ \hline
    Hybrid (Eqn \ref{eqn:hybrid}) & 59.3 & 62.1 & 68.9 & 77.1 & 79.9 & 81.7 \\ \hline
  \end{tabular}
\end{center}
\vspace*{-3mm}
  \caption{Comparison of model accuracies using a variety of samples sizes: accuracy uniformly increases with larger samples of the scores. See referenced equations (Eqn) for formal definitions of the models.} 
  \label{fig:contextlength}
  \vspace*{-4mm}
\end{table}

\subsection{Histogram Models}\label{subsec:hist}

The simplest models we consider are histogram models. Averaging the input data $\textbf{x}$ over voices and time gives us a histogram vector $h \in \{0,1\}^{N+D+1}$:
\begin{equation*}
h(\textbf{x}) = \frac{1}{TP}\sum_{t=1}^T\sum_{p=1}^P \textbf{x}_{t,p}.
\end{equation*}
Multiplying this histogram by a weight matrix $W_\theta \in \mathbb{R}^{(N+D+1)\times C}$ with parameterized entries gives us a simple linear model:
\begin{equation}\label{eqn:hist}
\vspace*{-1mm}
f_\theta(\textbf{x}) = W_\theta^\top h(\textbf{x}).
\end{equation}
No features are learned in this model; all that is learned are the linear weights $W_\theta$ on the histogram features. The model can be interpreted as a simplified version of the global feature models discussed in Section \ref{sec:rel}. In this case, the global features are the prevalences at which each of the $N+D+1$ note and duration symbols occur in a score.

\subsection{Voice Convolutional Models}\label{subsec:convpart}

Now let's consider a simple neural model inspired by $n$-gram features. Let $k$ be a number of features we desire to learn and $n$ be a number of time indices. Define the function $\text{relu} : \mathbb{R} \to \mathbb{R}$ by $t \mapsto t\textbf{1}(t > 0)$. Given a weight matrix $W_\theta^1 \in \mathbb{R}^{n(N+D+1) \times k}$ we can construct a ``convolutional'' feature representation $h_{t,p} \in \mathbb{R}^{T\times P \times k}$ at each time index $t$ in each voice $p$ defined by
\begin{equation}\label{eqn:conv1}
h_{t,p}(\textbf{x};\theta) = \text{relu}\left((W_\theta^1) ^\top \textbf{x}_{t:t+n,p}\right).
\end{equation}
We define $\textbf{x}_{t:t+n}$ to be a slice of $\textbf{x}$ from index $t$ to index $t+n$ (non-inclusive); when $t + n > T$, we pad $\textbf{x}$ with zeros. We then pool these features across voices and time to construct a single, global feature representation $h \in \mathbb{R}^k$, to which we can apply a linear classifier with weights $W_\theta \in \mathbb{R}^{k \times C}$:
\begin{align}\label{eqn:partconv}
\begin{split}
& h(\textbf{x};\theta) = \frac{1}{TP}\sum_{t=1}^T\sum_{p=1}^P h_{t,p}(\textbf{x};\theta),\\
& f_\theta(\textbf{x}) = (W_\theta)^\top h(\textbf{x};\theta).
\end{split}
\end{align}

This is a non-linear model (because of the non-linear \text{relu} ``activation'') and we can view $h$ as a learned feature representation of the score $\textbf{x}$. The weights (``filters'') $W_\theta^1$ learn to extract $k$ relevant patterns of length $n$ from voices, analogous to--but more expressive and compact than--classical $n$-gram featurizations. In our experiments we set $k = 500$ and $n=3$; the choice of $n$ is consistent with the pervasive use of 3-grams features in prior work \cite{hillewaere2010string,hontanilla2013modeling,wolkowicz2013evaluation,kempfert2018does}.

\subsection{Deeper Representations}\label{subsec:deep}

A natural way to extend the convolutional feature extraction discussed in Section \ref{subsec:convpart} is to compose multiple layers of convolutions. Given the feature representation $h_{t,p}$ given by Equation \ref{eqn:conv1} and a parameterized weight tensor $W_\theta^2 \in \mathbb{R}^{nk_1 \times k_2}$, we can construct a second layer of features
\begin{equation*}
h_{t,p}^2(\textbf{x};\theta) = \text{relu}\left((W_\theta^2)^\top h_{t:t+n,p}(\textbf{x};\theta)\right).
\end{equation*}
We can loosely interpret such a representation as building hierarchical features of features. In principle we can build arbitrarily deep stacks of features in this way; in our experiments, we were unable to realize significant gains using architectures with more than two convolutional layers.

Building a classifier over these features proceeds identically to the shallower models:
\begin{align}\label{eqn:partdeep}
\begin{split}
& h_\text{conv}(\textbf{x}; \theta) = \frac{1}{TP}\sum_{t=1}^T\sum_{p=1}^P h_{t,p}^2(\textbf{x};\theta) \\
& f_\theta(x) = (W_\theta)^\top h_\text{conv}(\textbf{x}; \theta).
\end{split}
\end{align}
For this model we set $n~=~3$, $k~=~300$, and $k_2~=~300$. 

\subsection{Full-Score Convolutional Models}\label{subsec:conv}

The models considered in Sections \ref{subsec:convpart} and \ref{subsec:deep} are largely monophonic: they extract features from individual voices (although they classify based on a pool of these features gathered from all the voices). Notably, those models have no ability to capture harmonic patterns in the interactions between voices. We now consider a model that can capture these interactions.

Let $W_\theta^1 \in \mathbb{R}^{nP(N+D+1) \times k}$, $W_\theta^2 \in \mathbb{R}^{nk \times k_2}$ and consider the model
\begin{align}\label{eqn:conv}
\begin{split}
& h_t(\textbf{x};\theta) = \text{relu}\left((W_\theta^1) ^\top \textbf{x}_{t:t+n}\right) \in \mathbb{R}^{T \times k},\\
& h_t^2(\textbf{x};\theta) = \text{relu}\left((W_\theta^1) ^\top h_{t:t+n}\right) \in \mathbb{R}^{T\times k2},\\
& h(\textbf{x};\theta) = \frac{1}{T}\sum_{t=1}^T h^2_t(\textbf{x};\theta),\\
& f_\theta(\textbf{x}) = (W_\theta)^\top h(\textbf{x};\theta).
\end{split}
\end{align}

We parameterize this model with $n = 3$, $k=300$ and $k_2=300$. This model is strictly more expressive than the part-wise models \eqref{eqn:partconv} or \eqref{eqn:partdeep}, capable of capturing patterns that the part models can't. However, the underperformance of this model \eqref{eqn:conv} relative to less expressive models \eqref{eqn:partconv} and \eqref{eqn:partdeep} suggest that it is prone to capture spurious patterns, leading to overfitting (compare results in Table~\ref{fig:overallresults}).

\subsection{Harmonic Models}\label{subsec:pitchconv}

All the models considered so far treat pitch classes as categorical data. We recognize, for example, that C4 is distinct from E4 or G4, but not that C4 is 4 semi-tones below E4 and 7 semi-tones below G4. This section introduces a model that exploits this structural order of pitch-classes, by convolving along the pitch-axis of the input tensor.

For notational convenience, we decompose the input tensor $\textbf{x} = \textbf{f} \oplus \textbf{d}$ into separate pitch components $\textbf{f} \in \{0,1\}^{T\times P \times N}$ and note-value components $\textbf{d} \in \{0,1\}^{T \times P \times (D+1)}$. Let $W_\theta^1 \in \mathbb{R}^{jP \times k}$ and convolve along the pitch-axis to construct a features $h_{t,n}(\textbf{f};\theta) \in \mathbb{R}^{T \times N \times k}$:
\vspace*{-2mm}
\begin{equation*}
h_{t,u}(\textbf{f};\theta) = \text{relu}\left((W_\theta^1)^\top\textbf{f}_{t,:,u:u+j}\right).
\end{equation*}
Here $j$ is a hyper-parameter indicating the height of the convolution; analogous to the width-$n$ hyperparameter in our time-domain convolutions for models \eqref{eqn:partconv}, \eqref{eqn:partdeep}, and \eqref{eqn:conv}. Unlike the time domain, we find that setting a large value of $j$ (in our models, $j = N/2$) is desirable; a similar phenomenon is observed for frequency-domain convolutions in \cite{thickstun2018invariances}.

We proceed to pool the features $h_{t,u}$ together across the pitch domain to construct $h_t \in \mathbb{R}^{T \times k}$:
\vspace*{-2mm}
\begin{equation}\label{eqn:pitchconv}
h_t(\textbf{f};\theta) = \frac{1}{N}\sum_{u=1}^N h_{t,u}(\textbf{f},\theta).
\end{equation}
The idea of this pooling is to construct a feature-set that is invariant to pitch translation. We are interested in learning features such as, for example, the occurrence of general major chords rather than the occurrence of a particular major chord such as the one rooted at A3. The pooling operation above precludes us from learning the latter type of feature.

We then construct a second layer of features to integrate the harmonic features $h_t$ together with the note-value features $\textbf{d}_t$. Using weights $W_\theta^2 \in \mathbb{R}^{k \times k_2}$ and $W_\theta^3 \in \mathbb{R}^{(D+1) \times k2}$ we build $h_t^2(\textbf{x};\theta) \in \mathbb{R}^{T \times k_2}$. We then pool the representations $h_t^2$ across time and construct a linear classifier on the resulting representation:

\vspace*{-2mm}
\begin{align}\label{eqn:harmony}
\begin{split}
& h_t^2(\textbf{x};\theta) = \text{relu}\left((W_\theta^2)^\top h_t(\textbf{f};\theta) + (W_\theta^3)^\top \textbf{d}_t\right),\\
& h_\text{harmonic}(\textbf{x};\theta) = \frac{1}{T}\sum_{t=1}^T h^2_t(\textbf{x};\theta),\\
& f_\theta(\textbf{x}) = (W_\theta)^\top h_\text{harmonic}(\textbf{x};\theta).
\end{split}
\end{align}

We parameterize this model with $k=64$ and $k_2=500$. 

\subsection{Hybrid Models}\label{subsec:hybrid}

Looking back at the models we've introduced, observe that the voice models \eqref{eqn:partconv} and \eqref{eqn:partdeep} exploit temporal structure within voices, but pool away any harmonic patterns between voices. In contrast, the harmonic model \eqref{eqn:harmony} exploits harmony between voices but pools away any sequential patterns across time indices. The full-score convolutional model can capture both types of structure, but is prone to capture spurious patterns and overfit.

This motivates the introduction of our final, hybrid model that weakly combines temporal and harmonic models to increase predictive power without overfitting. The idea is to feed the input tensor separately through temporal and harmonic models to construct features representations $h_\text{conv}$ \eqref{eqn:partdeep} and $h_\text{harmonic}$ \eqref{eqn:harmony} respectively. We combine these features in a final, linear layer using weights $W_\theta^c \in \mathbb{R}^{k_2 \times C}$ and $W_\theta^h \in \mathbb{R}^{k_2 \times C}$ to make a prediction:
\begin{align}\label{eqn:hybrid}
\begin{split}
& f_\theta(\textbf{x}) = (W_\theta^\text{c})^\top h_\text{conv}(\textbf{x};\theta) + (W_\theta^\text{h})^\top h_\text{harmonic}(\textbf{x};\theta).
\end{split}
\end{align}
Because temporal and harmonic information are only combined in the final linear layer, this model is unable to learn expressive relationships between these features, such as the classical XOR relationship \cite{minsky2017perceptrons}. As we see in Table \ref{fig:overallresults}, this combination increases accuracy over either the temporal or harmonic models on their own.

\vspace{-3mm}
\section{Results and Conclusions}\label{sec:results}

The results of all models discussed in this paper, evaluated on the full corpus, are presented in Table \ref{fig:overallresults}. We sort the rows in this table by the number of scores for each composer; we observe a trend towards increasing accuracy when we have more data (with some outliers).

{\tiny
\begin{table}[h!]
  \centering
  \begin{tabular}{lrrrrrr}
    \toprule
    \cmidrule{1-7}
    & \multicolumn{6}{ c }{Models}\\
    \midrule
    Composer & \eqref{eqn:hist} & \eqref{eqn:partconv} & \eqref{eqn:partdeep} & \eqref{eqn:conv} & \eqref{eqn:harmony} & \eqref{eqn:hybrid} \\
    \midrule
    \midrule
    Japart  & 0.0 & 13.6 & 13.6 & 9.1 & 18.2 & 13.6 \\
    Hummel  & 41.7 & 54.2 & 66.7 & 62.5 & 87.5 & 91.7 \\
    Compere  & 0.0 & 25.9 & 22.2 & 25.9 & 40.7 & 37.0 \\
    Vivaldi  & 30.3 & 94.4 & 91.6 & 54.5 & 45.5 & 54.5 \\
    Du Fay & 45.7 & 82.9 & 74.3 & 71.4 & 80.0 & 74.3 \\
    Orto  & 0.0 & 18.6 & 37.2 & 25.6 & 46.5 & 48.8 \\
    Joplin & 85.1 & 91.5 & 93.6 & 93.6 & 95.7 & 91.5 \\
    D. Scarlatti  & 44.1 & 59.3 & 62.7 & 78.0 & 79.7 & 72.9 \\
    Busnois  & 13.2 & 48.5 & 48.5 & 51.5 & 60.3 & 60.3 \\
    Chopin & 55.3 & 54.2 & 64.5 & 72.4 & 76.3 & 68.4 \\
    Ockeghem & 13.3 & 55.1 & 69.4 & 52.0 & 66.3 & 72.4 \\
    Martini  & 44.3 & 68.0 & 75.4 & 59.8 & 68.0 & 73.8 \\
    Mozart & 34.8 & 56.3 & 61.6 & 63.6 & 70.2 & 67.5 \\
    Beethoven  & 72.2 & 82.2 & 83.4 & 78.7 & 84.0 & 89.3 \\
    de la Rue  & 27.5 & 57.9 & 71.3 & 63.5 & 73.6 & 79.2 \\
    Corelli  & 89.4 & 89.9 & 86.2 & 93.1 & 93.6 & 95.2 \\
    Haydn & 85.6 & 75.6 & 71.3 & 79.9 & 82.3 & 83.7 \\
    Josquin  & 81.1 & 78.7 & 76.4 & 75.9 & 77.3 & 82.3 \\
    Bach & 92.3 & 95.7 & 96.1 & 97.2 & 97.2 & 97.6 \\
    \midrule
    Overall & 64.2 & 75.4 & 76.9 & 75.5 & 79.8 & 81.7 \\
    \bottomrule
  \end{tabular}
  \caption{Results of the 19-way classification problem on the full corpus for each model considered in this paper. } 
 \label{fig:overallresults}
 \vspace*{-3mm}
\end{table}
}

To compare with previous work, we train additional models on subsets of the corpus. We invite comparisons between the results in Table \ref{fig:renaissanceresults} and the results of \cite{brinkman2016musical}, and between the results in Table \ref{fig:classicalresults} and the results of \cite{herremans2016composer}. These comparisons are imperfect: neither \cite{brinkman2016musical} nor \cite{herremans2016composer} report the precise scores used in their experiments. Nevertheless our corpus is derived from the same KernScores sources as \cite{brinkman2016musical} and \cite{herremans2016composer}, and contains a comparable number of scores to the counts reported in \cite{herremans2016composer}. Therefore we believe our subsets are similar to the corpora used in these works and that comparison is meaningful. For future reference, the exact dataset used for the present work can be found online.\footnote{{\color{blue}\url{http://homes.cs.washington.edu/~thickstn/ismir2019classification/}}}

\begin{table}[t]
\begin{center}
\setlength\tabcolsep{5pt}
  \begin{tabular}{ | l || c | c | c | c | c | c|}
    \hline
    & Bach & Orto & Fay & Ock. & Josq. & Rue\\ \hline\hline
    Bach & 100.0 & 0.0 & 0.0 & 0.0 & 0.0 & 0.0 \\ \hline
    Orto & 0.0 & 39.5 & 0.0 & 7.0 & 51.2 & 2.3 \\ \hline
    Du Fay & 0.0 & 0.0 & 82.9 & 11.4 & 5.7 & 0.0 \\ \hline
    Ockegham & 0.0 & 2.0 & 5.1 & 81.6 & 9.2 & 2.0 \\ \hline
    Josquin & 0.7 & 1.4 & 1.2 & 3.3 & 84.4 & 9.0 \\ \hline
    de la Rue & 1.1 & 0.0 & 0.0 & 0.6 & 25.8 & 72.5\\
    \hline
  \end{tabular}
  
  \vspace{3mm}
  
  \begin{tabular}{ | c || c | c | c | c | c | c|}
   
    \hline
    \hspace{15mm} & Bach & Orto & Fay & Ock. & Josq. & Rue\\ \hline\hline
    \eqref{eqn:hybrid} & \textbf{100.0} & \textbf{39.5} & \textbf{82.9} & \textbf{81.6} & \textbf{84.4} & 72.5 \\ \hline
    KNN\cite{brinkman2016musical} & 94.5 & 38.9 & 42.9 & 70.0 & 60.6 & 80.6 \\ \hline
    SVM\cite{brinkman2016musical} & 98.5 & 33.3 & 25.0 & 60.0 & 60.0 & \textbf{87.1} \\ \hline
  \end{tabular}
\end{center}
\vspace*{-3mm}
  \caption{(Top) Confusion matrix for the hybrid model \eqref{eqn:hybrid}, trained and evaluated on a 6-composer subset of the corpus. Compare to the results in Tables 3 and 4 (page 6) of \cite{brinkman2016musical}. (Bottom) Accuracy comparisons of our hybrid model to the KNN and SVM models from \cite{brinkman2016musical}. }
  \label{fig:renaissanceresults}
\end{table}

\begin{table}
\begin{center}
  \begin{tabular}{ | c || c | c | c |}
    \hline
    & Bach & Haydn & Beethoven \\ \hline\hline
    Bach & 99.8 & 0.2 & 0.0 \\ \hline
    Haydn & 3.4 & 93.3 & 3.3 \\ \hline
    Beethoven & 3.0 & 10.6 & 86.4 \\
    \hline
  \end{tabular}
  
  \vspace{3mm}
  
  \begin{tabular}{ | c || c | c | c |}
    \hline
    & Bach & Haydn & Beethoven \\ \hline\hline
    \eqref{eqn:hybrid} & \textbf{99.8} & \textbf{93.3} & \textbf{86.4} \\ \hline
    SVM \cite{herremans2016composer} & 94.6 & 80.3 & 64.8 \\ 
    \hline
  \end{tabular}
\end{center}
  \caption{(Top) Confusion matrix for the hybrid model \eqref{eqn:hybrid}, trained and evaluated on a 3-composer subset of the corpus. Compare to the results in Table 9 (page 18) of \cite{herremans2016composer}. (Bottom) Accuracy comparisons of our hybrid model to the SVM model from \cite{herremans2016composer}.} 
  \label{fig:classicalresults}
  \vspace{-3mm}
\end{table}

For the popular Haydn versus Mozart string quartet classification task \cite{hillewaere2010string,herlands2014machine,van2005musical,kempfert2018does}, we were unsuccessful. The standard evaluation metric for this task is LOOCV, which we could not perform due to the computational expense of our models. With 10-fold cross validation, we observed exceedingly high variance upon repeat optimizations of the same model. However none of our optimizations exceeded 80\%. Due to imbalance between Haydn and Mozart quartets (209 versus 82 scores) a classifier that simply predicts Haydn given any input achieves 71.8\%.

Overall, we conclude that the convolutional models proposed in this paper perform quite well. We find this notable, given that success in neural modeling is often associated with much larger datasets. Furthermore, we do not believe that the potential of these methods has been exhausted; further investigation may yield even better convolutional architectures for composer classification.

\section{Acknowledgements}

This work was supported by NSF Grant DGE-1256082. We also thank NVIDIA for their donation of a GPU.

\bibliography{references}

%
%
%
%

\end{document}